\documentclass[runningheads]{llncs}
\usepackage{graphicx}
\begin{document}
\graphicspath{ {./images/} }
\title{Sentiment and Knowledge Based Algorithmic Trading with Deep Reinforcement Learning}
\titlerunning{Trading with Deep Reinforcement Learning}
\author{Abhishek Nan\thanks{Equal Contribution}\inst{1} \and
Anandh Perumal\textsuperscript{*}\inst{1,2} \and
Osmar R. Zaiane\inst{1,2}}
\authorrunning{A. Nan et al.}
\institute{University of Alberta, Canada \and
Alberta Machine Intelligence Institute\\
\email{\{anan1,anandhpe,zaiane\}@ualberta.ca}}
\maketitle              % typeset the header of the contribution

\begin{abstract}
Algorithmic trading, due to its inherent nature, is a difficult problem to tackle; there are too many variables involved in the real world which make it almost impossible to have reliable algorithms for automated stock trading. The lack of reliable labelled data that considers physical and physiological factors that dictate the ups and downs of the market, has hindered the supervised learning attempts for dependable predictions. %Here, t
To learn a good policy for trading, we formulate an approach using reinforcement learning which  uses traditional time series stock price data and combines it with news headline sentiments, while leveraging knowledge graphs for exploiting news about implicit relationships.

\keywords{Reinforcement Learning  \and Trading \and Stock Price Prediction \and Sentiment Analysis \and Knowledge Graph.}
\end{abstract}
\section{Introduction}
Machine learning is mainly about building predictive models from data. When the data are time series, models can also forecast sequences or outcomes. Predicting how the stock market will perform is an application where people have naturally attempted machine learning but it turned out to be very difficult because involved in the prediction are many factors, some rational and some appearing irrational.  Machine learning has been used in the financial market since the 1980s~\cite{doi:10.1111/j.1540-5915.1987.tb01533.x}, trying to predict future returns of financial assets using supervised learning such as artificial neural networks~\cite{atsalakis2009}, support vector machines~\cite{karazmodeh2013} or even decision trees~\cite{panigrahi15}; but so far, there has been only limited success. There are multiple causes for this. For instance, in supervised machine learning, we usually have labelled datasets with balanced class distributions. When it comes to the stock market, there is no such labelled data for when someone should have bought/sold their holdings. This leads credence to the problem being fit for the reinforcement learning framework~\cite{sutton1998introduction}, a behavioral-based learning paradigm relying on trial and error and supplemented with a reward mechanism. Reinforcement learning has the ability to generate this missing \textit{labelling} once we define a proper reward signal. But there are still other issues in this context which are specific to stock markets. They are prone to very frequent changes and often these changes cannot be inferred from the historical trend alone. They are affected by real world factors such as political, social and even environmental factors. For instance, an earthquake destroying a data-center could result in stock prices dropping for a company; a new legislation about trade can positively impact the value of a company. Noise to signal ratio is very high in such conditions and it becomes difficult to learn anything meaningful under such circumstances. Such environments can be modelled as Partially Observable Markov Decision Processes (POMDPs) ~\cite{ASTROM1965174}, where the agent only has limited visibility of all environmental conditions. A POMDP models an agent decision process in which it is assumed that the system dynamics are determined by an a discrete time stochastic control process, but the agent cannot directly observe the underlying state~\cite{Kaelbling98}.
Our contribution is the use of sentiment analysis done on news related to a traded company and its services in conjunction with a reinforcement algorithm to learn an appropriate policy to trade stocks of the given company. To find the relevant news title on-which to apply sentiment analysis, we use a traversal of a knowledge graph. 

After highlighting the related work in Section \ref{sec:relatedwork}, we present our approach in Section \ref{sec:headings} combining headlines from news and their sentiment after finding their relation with the relevant stock hinging on a knowledge graph, and finally learning a good policy for buying and selling using Reinforcement Learning. We present an empirical evaluation using the stocks of Microsoft between 2014 and 2018 in Section \ref{sec:evaluation}. Section \ref{sec:results} highlights the analysis of the observed results. Finally, we conclude and present perspective on future work in Section \ref{sec:conclusion}.

\section{Related Work}\label{sec:relatedwork}
There have been many approaches in the past which try to model traditional time series approaches for stock price prediction~\cite{sharma14,taran15,Jain18}. The main idea in these approaches is to predict the stock price at the next time step given the past trend. This prediction is then fed to a classifier which tries to predict the final buy/sell/hold action. Most modern deep learning techniques try to use some form of recurrent networks to model the sequential trend in the data. \cite{7364089} used LSTMs with great success to make predictions in the Chinese stock market. Approaches integrating some form of event data has been explored as well to some extent. For instance, \cite{doi:10.1002/(SICI)1099-1174(199703)6:1<11::AID-ISAF115>3.0.CO;2-3} used manually extracted features from news headlines to integrate event information and spliced them with several other economic indicators according to prior knowledge and combined them together as the input to neural networks.

An alternative approach is to use Reinforcement Learning. \cite{Fischer18} shows in a comprehensive survey on the use of RL in financial markets that there are many attempted approaches but the problem is far from being solved \cite{Fischer18}. From a reinforcement learning (RL) perspective, \cite{TAN20114741} proposed an Adaptive Network Fuzzy Inference System (ANFIS) supplemented by the use of RL as a non-arbitrage algorithmic trading system. \cite{7407387} use a deep learning component which automatically senses the changing market dynamics for feature learning and these features are used as input to an RL module which learns to make trading decisions. \cite{tomgrek} explored the use of actor-critic methods for stock trading and serves as one of the primary motivations behind our research.

Furthermore, public opinion can often provide valuable indication as to how a company might be posed to perform in the market. Attempts have been made previously to directly classify each comment on a stock trading forum as a indicator of a buy/sell/hold decision \cite{sprenger2014tweets}. Rather than use text data in its entirety as a variable to make decisions, the general sentiment of the text can be extracted as a score \cite{doi:10.1111/j.1468-5957.2011.02258.x} and combined with other related data.

Our approach in a way tries to take the best of these methods and extend them into a single dynamic system paired with knowledge graphs. We extract sentiments from event information and use knowledge graphs to detect implicit relationships between event information and a given traded company. We then combine this information with the time series stock data, and allow our agent to learn an optimal policy using deep reinforcement learning. We also take advantage of more recent deep RL techniques such as the DQN introduced by \cite{mnih2015human}. %(\citeyear{mnih2015human}). 
This approach of combining knowledge graph driven sentiment data with deep RL is our novel proposal and has not been explored in any literature we surveyed.

\section{Proposed Approach}
\label{sec:headings}
Our approach combines concepts from a few different domains; hence, we give a short overview of each of them and %then 
connect how they are used in our project.

\subsubsection{Q-learning}
Q-learning is a model free reinforcement learning algorithm. Given an environment, the agent tries to learn a policy %(a way of behaviour) 
which maximises the total reward it gets from the environment at the end of an episode (a sequence of interactions). For instance, in our problem setting, an episode during training would be the sequence of interactions the agent makes with the stock market starting from January 1, 2014 and ending on December 31, 2017. The agent would try to learn a behaviour which maximises the value of its portfolio at the end date.

The intuition behind Q-learning is that the agent tries to learn the utility of being in a certain state and taking a particular action in that state and then following the behavioural policy learnt so far till the end of the episode (called the action value of that state). So, Q-learning tries to learn the action value of every state and action. It does this by exploring and exploiting at the same time. For instance, %say 
a trading agent starts on Day 1 and it has two options: \textit{Buy} and \textit{Sell}. It takes the \textit{Buy} option (say arbitrarily) the first time it experiences Day 1 and receives a reward of 10 units. For optimal performance, the agent will \textit{usually} follow the best possible option available to it. \textit{Usually}; because if it always followed the best option that it \textit{thinks} is available to it, it won't learn the value of taking the other options available to it in that state. For instance, in the above example, \textit{Sell} could have led to a reward of 20, but it would have never known this if it always took the \textit{Buy} option after it first experienced Day 1 with a \textit{Buy} action. This dilemma is known as the exploration and exploitation trade-off. A naive, yet effective way of solving this is to always take the "greedy" option, except also act randomly a small percentage of the time, say with a probability of 0.1. This is known as the $\epsilon$-greedy approach with $\epsilon=0.1$. This finally brings us to the Q-learning equation, which updates the action values of each state and action pair.

\begin{large}
$Q(S_t,a_t) \leftarrow Q(S_t,a_t) + $$\alpha[R_{t+1} + \gamma \max\limits_{a} Q(S_{t+1},a)-Q(S_t,a_t)]$
\end{large}

\subsubsection{Function approximation}
A shortcoming of the above mentioned Q-learning methodology is the obvious fact that it relies on the idea of a distinct state. What this means is that the Q-learning update can only be applied to an environment where each state(\textit{$s_t$}) can be distinctly labelled. This would mean we would have to maintain a huge table of every possible state and action combination that can be encountered and their action values. This does not generalize very well and is not tractable for real world problems. For instance, given today's state of the world to a stock trading agent, it might make some decision and learn from it, but it's very unlikely that the exact same conditions will ever be presented to it again. The solution to this is to use a function approximator, which given the current environmental observation and the chosen action maps them to an action value. The parameters of the approximator can then be updated similar to supervised learning once we have observed the actual reward. In our experiments, we use an artificial neural network for function approximation.

For large state spaces since optimizing artificial neural networks via just back-propagation becomes unstable so we adapt the modifications to a Deep Q-Network (DQN) as presented by \cite{mnih2015human}. These modifications include gradient clipping, experience replay and using a Q-network which periodically updates an independent target network.

\subsection{Sentiment analysis}
Sentiment analysis is an automated process to annotate text predicted to be expressing a positive or negative opinion.
Also known as opinion mining, sentiment analysis categorizes text into typically two classes positive vs. negative, and often a third class: neutral.
Discovering the polarity of a text is often used to analyze product or service reviews, like restaurants, movies, electronics, etc. but also other written text like blog posts, memos, etc. There are two main types of sentiment analysis approaches, namely lexicon-based using a dictionary of words with their polarities; and machine learning based which build a predictive model using a labelled train dataset~\cite{yaddolahi17}.   

Each sentence/sequence of sentences in a language in general, has a positive or a negative connotation associated with it; sometimes neutral. A news headline, the full news article itself, or even this paper, typically express an opinion to some degree. Natural Language Processing techniques are used to extract such connotations in an automated manner \cite{godbole2007large}. Once extracted, %this allows the use of machine learning techniques in domains where opinions 
it can serve as a vital data point for applications such as in marketing to understand customers' opinion, as mentioned above. In our case we would like to use sentiment analysis to assess whether a news headline is favorable or admonitory to the company for which we are trading stocks.

Consequently, in our case, each news headline is posited to be either positive, negative or neutral from the perspective of the company we are considering trading stocks for. Positive sentiments can predict a general upturn in stock prices for a company, and similarly negative sentiments can possibly indicate a downturn \cite{fisher2000investor,si2013exploiting}. %While sentiments can be directly extracted from any text corpus (news headlines in this case), a lot more implicit information can be obtained by pairing knowledge graphs with this approach.

\subsection{Knowledge Graphs}
Lexical thesauri and ontologies are databases of terms interconnected with semantic relationships. %Some examples include WordNet\footnote{https://wordnet.princeton.edu/} for English terms and the Unified Medical Language System (UMLS)\footnote{https://www.nlm.nih.gov/research/umls/} for terms in the medical domain. 
They are often represented in a graph with entities and relationships. A knowledge base or a knowledge graph, are more complex graphs where the entities are not simple terms but a composite of knowledge. %Some examples include DBpedia\footnote{https://wiki.dbpedia.org/} or Google Knowledge Graph, which we use in this work.

The Google Knowledge Graph was specifically created to enhance the results of a Google search. Traditionally a Web search used to be limited to string matching keywords in an entire corpora to a given query. However, since entities in the real world are linked to each other and this link can be expressed in different ways, simple string matching is not adequate for an intelligent search. This interconnection is characterized in the knowledge graph which represents a graph-like data structure where each node is an entity and the edges between the nodes indicate the relationships between them. %The most usage we see is probably Google's version of it implemented in the Google Search Engine. 
For instance, a naive search for ``Bill Gates" using simple string matching would not bring up Microsoft. However, with a knowledge graph, since ``Bill Gates", being the principal founder of Microsoft, he is a very relevant node close to the "Microsoft" node in a knowledge graph and hence, ``Microsoft" would be brought up as a relevant search result. This way entities which are related to a company, but not explicitly mentioned in the news headline, can be identified as potential factors impacting stock prices. In our case, headlines covering Excel, Windows, Azure, Steve Ballmer, or Satya Nadella, or other entities connected to Microsoft in the knowledge graph, would be passed to the sentiment analysis and their polarity exploited in the learning algorithm.     

\section{Empirical Evaluation} \label{sec:evaluation}

\subsection{Data}

\textbf{Stock data:}
We used stock data from the Yahoo Finance API\footnote{https://finance.yahoo.com/quote/MSFT/history/} dated from January 1, 2014 to December 31, 2017 for our training environment. The data for the test period is from January 1, 2018 to December 31, 2018. In our experiment, for both training and testing cases we used Microsoft Corporation's (MSFT) stock data - i.e. we trained our agent to trade Microsoft stocks.

\textbf{Sentiment Data:}
For news information, we scraped historical news headlines from the Reuters Twitter account\footnote{https://twitter.com/reuters} using a python scraper\cite{taspinar}. The time period of the news headlines corresponds exactly to the stock data, i.e. training data from January 1, 2014 to December 31, 2017 and testing data from January 1, 2018 to December 31, 2018.

Next, for each news headline we remove stopwords and tokenize it. Each token is then checked for the existence of an \textit{Organization} node relationship with the specific company of interest (Microsoft Corporation in our case) in a knowledge graph within a pre-specified distance. In our experiment we chose a distance measure of 5. Selecting a walk-length longer than this resulted in too much noise, and any shorter meant there would be very few implicit relationships found. This value was tuned empirically on the basis of some manual experiments we performed. For our experiments, we used the Google Knowledge Graph\footnote{https://developers.google.com/knowledge-graph/}. Once we find that any token in a headline is within this pre-specified distance of our organization (Microsoft), by extension we deem the entire headline as relevant to the organization in consideration. This is a naive approach, but allows us to make better use of news data that might not be directly linked to Microsoft, but might have indirect consequences. For instance, a news headline talking about Azure, which is Microsoft's cloud service offering, would not get identified as a news affecting MSFT stock prices, but by using a knowledge graph, we can uncover this implicit relationship.

Once we have headlines relevant to Microsoft, we use an ensemble sentiment analyser for sentiment classification. Since some headlines proved to be tricky to classify correctly by any single available sentiment classifier, we tried this approach of using an ensemble comprising of IBM Watson~\cite{ibmwatson}, TextBlob~\cite{textblob} and NLTK~\cite{nltk}. We classify each news headline as positive and negative news and use the classification from the classifiers above, choosing whichever one has the highest confidence. If there are multiple headlines on the same day, we use the majority of the sentiment score from all headlines for that day leading to a net +1 if majority is positive sentiment and -1 if majority is negative sentiment. 
An example positive headline dated 2016-07-14 can read: ``Microsoft wins landmark appeal over seizure of foreign emails.", while an example of a headline expressing a negative sentiment dated 2015-12-31 is ``Former employees say Microsoft didn't tell victims about hacking."

\subsection{MDP Formulation}

\textbf{Episode:}
A single episode consists of the agent interacting with the stock trading environment once per day starting from Januray 1, 2014 and lasts till December 31, 2017 (for the training period). The agent explores different policies and improves its existing policy as more and more episodes elapse.

\begin{figure}[!tbp]
\caption{Construction of agent's state}
\centering
\includegraphics[width=1.2\textwidth]{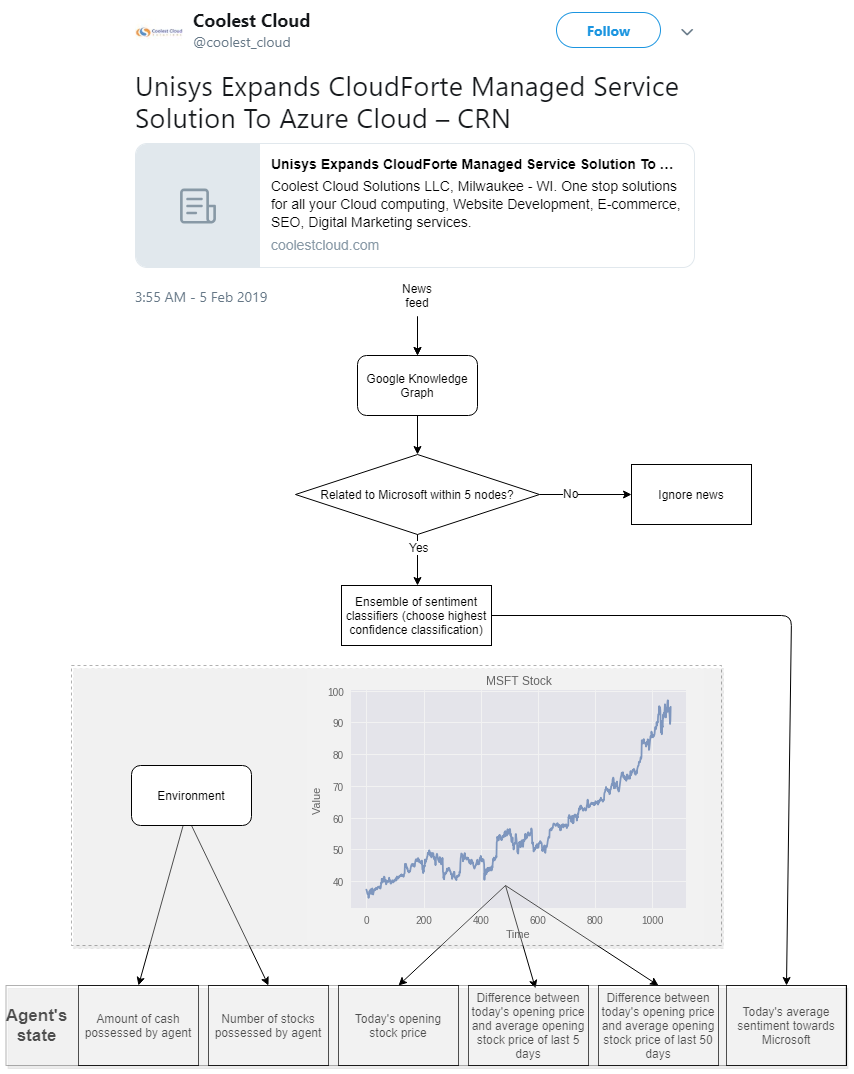}
\label{fig:QStockTrader}
\end{figure}

\textbf{State:}
Our current environment describes each state using 6 variables:% (Figure~\ref{fig:QStockTrader}):
\begin{enumerate}
    \item Current amount of money the agent has;
    \item Current number of stocks the agent has;
    \item Opening stock price on today's date;
    \item Difference between today's opening price and average opening price of last 5 days' window;
    \item Difference between today's opening price and average opening price of last 50 days' window;
    \item Average sentiment towards the company for today's date.
\end{enumerate}
While (1), (2), (3) are values necessary for maintaining the state of the agent, (4) and (5) were added to give it some indication of the trend in the stock prices. (4) provides the trend over a short time window (5 days), while (5) provides the trend information over a longer time window (50 days). (6) provides the sentiment information calculated as described in the previous section. In short, relevance of headlines are assessed with a knowledge graph. The sentiment expressed in the relevant pieces are used in (6). Figure \ref{fig:QStockTrader} shows the entire workflow for the construction of the agent's state before it goes into the DQN.

\textbf{Action space:}
The agent, our stock trading bot, interacts with this environment on a per day basis. It has the option to take three actions: (1) Buy a stock; (2) Sell a stock; and (3) Do nothing/Hold.
%\begin{enumerate}
%    \item Buy a stock
%    \item Sell a stock
%    \item Do nothing/Hold
%\end{enumerate}

\textbf{Rewards:}
The intuition behind rewards is to provide a feedback signal to the agent to allow it to learn which actions are good/bad based on when they were taken. So, in our case, a net increase in portfolio at the end of the trading period should lead to a positive reward, while a net loss would lead to a negative reward. So, our initial attempts focused on this reward scheme where the agent's reward was the net profit/loss after 3 years (2014-2017). But, this reward signal proved too sparse to train on, since the agent got just one single reward after 3 years of activity and it is difficult for it to know which action taken when (over 3 years) contributed to the final reward. The agent just learnt to "Do nothing", since as the result of a general increasing trend in the MSFT stock price, it led to a small net increase in the portfolio and this was a local optima for the agent which it could not move out of due to the sparse reward signals.

Finally, after plenty of experimentation with the reward scheme, we arrived at one where it was rewarded for not just making a profit, but also for buying/selling on a day to day basis. If on any given day, it decided to Buy or Sell, it was given a reward of +1 for making a profit and -1 for making a loss. It was given a small negative reward of -0.1 for "Doing nothing" to discourage it from being passive for extended periods. A reward of -10 was given if it ran out of money, but still had stocks. A reward of -100 was given in case it went completely bankrupt with 0 stocks in hand and no money to buy a single stock.

\textbf{Deep Q-Network (DQN) Architecture:}
The DQN used two identical neural networks (Q-network and target network) each with 3 hidden layers for function approximation. Each hidden layer had a size of 64 units and used ReLU activation. The input layer had 6 input nodes corresponding to each state feature. The output layer had 3 nodes corresponding to the action space. The experience replay buffer size was restricted to a size of 1000. 

\textbf{Training:}
The DQN was trained with mini-batch gradient descent using Adam\cite{kingma2014adam} on the Huber loss. During training the agent started off with \$1000 USD and 10 MSFT shares on January 1, 2014 and interacted with the environment till December 31, 2017. The agent in the form of a DQN is trained over 2000 epochs. 

% Figure \ref{fig:training_graph}.A presents how the return (sum of rewards for all actions taken as specified by the reward scheme). The agent initially starts with a large negative value (representing a high loss portfolio) and then 
% gradually converges towards a better policy (which possibly yields profits) as more training episodes elapse.

% \begin{figure}[!htbp]
% \caption{A: Training score vs the number of episodes for the agent with sentiment information; B: Total portfolio value over testing period (January 1, 2018 to December 31, 2018). }
% \centering
% \includegraphics[width=0.49\textwidth]{training_graph}
% \includegraphics[width=0.49\textwidth]{test_portfolio}
% A \hspace{2in} B
% \label{fig:training_graph}
% \end{figure}

\begin{figure}[!htbp]
\caption{\textbf{Performance of different agents for different stocks} (Left column: Train data; Right column: Test data)}
\centering
\includegraphics[width=1.1\textwidth]{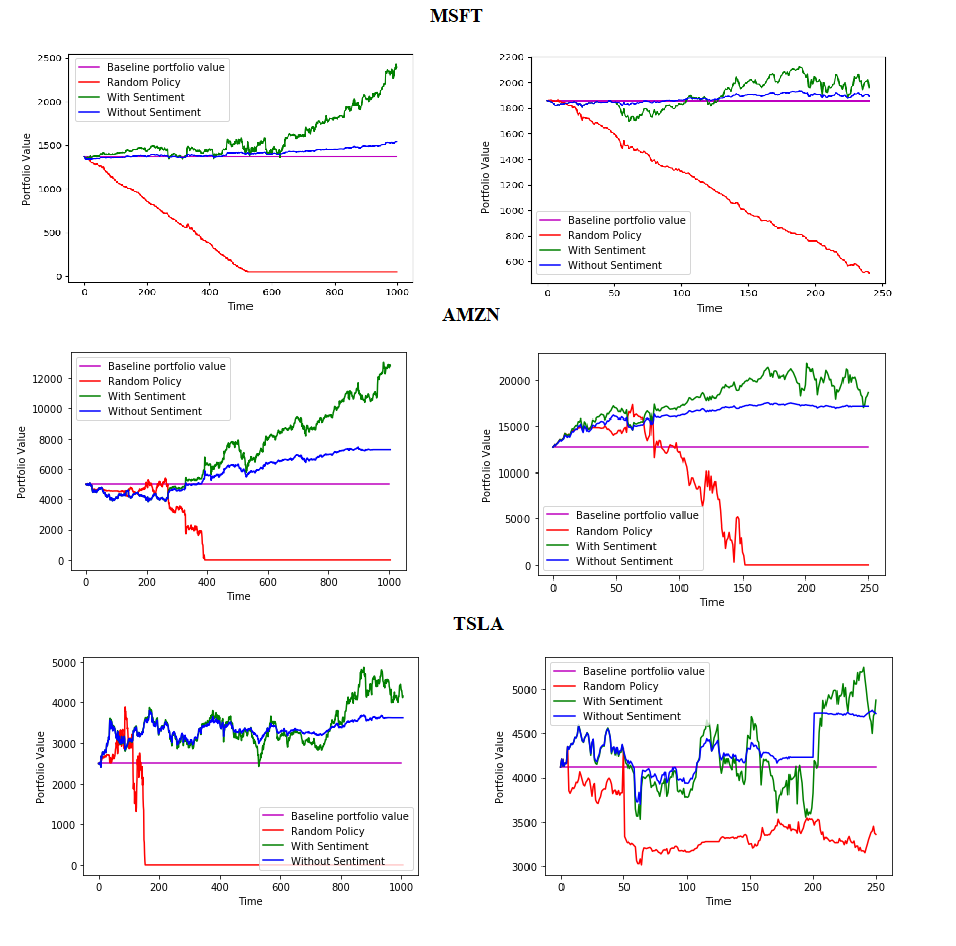}

%\includegraphics[width=0.50\textwidth]{training_portfolio}
%\includegraphics[width=0.49\textwidth]{images/test_portfolio.png}

%\includegraphics[width=0.49\textwidth]{images/amazon_training.png}
%\includegraphics[width=0.49\textwidth]{images/amzn_test.png}

%\includegraphics[width=0.49\textwidth]{images/TSLA_train.png}
%\includegraphics[width=0.49\textwidth]{images/TSLA_train.png}
%\label{fig:Google_training_portfolio}
\label{fig:training_portfolio}
\end{figure}

\section{Results and Analysis} \label{sec:results}
The primary hypothesis of this work was that providing sentiment information to the agent on a daily basis would add to its performance ceiling and it would be able to make more profit via trading. Therefore, we compare both approaches, i.e. an agent with sentiment data provided and another agent without any sentiment data provided. We evaluate both on our test data set, which spans from January 1, 2018 to December 31, 2018. If sentiments do add any additional value to the environment, it should be able to make more profit.

\subsection{Training data analysis}
Before looking at the performance on the test data, we also analyse the performance of both models on the training data as well. Figure \ref{fig:training_portfolio} shows this comparative analysis. The \textit{Baseline Portfolio Value} is the starting portfolio value of the agent (i.e. the net value of 1000\$ and 10 stocks on starting day). The \textit{Random Policy} is an agent which takes random actions (Buy, Sell, Hold) on each interaction. As expected, a random policy agent goes broke soon enough and makes no profit. The agent with no sentiment input, does learn a policy good enough to make profit, but nowhere near good enough as compared to the agent which had sentiment input.

%\begin{figure}[!htbp]
%\caption{\textbf{Total portfolio value of over training period} (Microsoft: January 1, 2014 to %December 31, 2017)}
%\centering
%\includegraphics[scale=0.55]{training_portfolio}
%\label{fig:training_portfolio}
%\end{figure}

Figure \ref{fig:training_portfolio} shows the same trend with the same training done on data about Microsoft, Amazon and Tesla stocks,. We can distinctly see that the learned policy using sentiment from news headlines outperforms the policy that only considers stock data.

%\begin{figure}[!htbp]
%\caption{\textbf{Performance of different agents for different stocks} (Left column: Train %data; Right column: Test data)}
%\centering
%\includegraphics[width=1.1\textwidth]{images/All_Stocks.png}
%
%%\includegraphics[width=0.50\textwidth]{training_portfolio}
%%\includegraphics[width=0.49\textwidth]{images/test_portfolio.png}
%
%%\includegraphics[width=0.49\textwidth]{images/amazon_training.png}
%%\includegraphics[width=0.49\textwidth]{images/amzn_test.png}
%
%%\includegraphics[width=0.49\textwidth]{images/TSLA_train.png}
%%\includegraphics[width=0.49\textwidth]{images/TSLA_train.png}
%%\label{fig:Google_training_portfolio}
%\label{fig:training_portfolio}
%\end{figure}
%\vspace{-0.2in}
\subsection{Test data analysis}
For Microsoft, during the test period, the stock prices at the beginning start quite a bit higher (approx. \$85 in January 2018) as compared to the training period start (approx. \$40 in January 2014). Despite this, the MSFT agent (Figure \ref{fig:training_portfolio}) learned a policy good enough to generate profit, both with and without sentiment data. However, in general, it was not able to generalize to the test data as well as we saw during training, and its profits dropped. But still, the agent with sentiment information ends up making more profit than the agent without sentiment data. Similar trends are present for both the other stocks as well.
%\begin{figure}[!htbp]
%\caption{\textbf{Total portfolio value over testing period (January 1, 2018 to December 31, %2018)} left: Google; Right: Microsoft:}
%\centering
%%\includegraphics[scale=0.55]{test_portfolio}
%\includegraphics[width=0.45\textwidth]{googlestock}
%\label{fig:test_portfolio}
%\end{figure}

\subsection{Sharpe ratio}
The Sharpe ratio is another measure that is often used in trading as a means of evaluating the risk adjusted return on investment. It can be used as a metric to evaluate the performance of different trading strategies. It is calculated as the expected return of a portfolio minus the risk-free rate of return, divided by the standard deviation of the portfolio investment. In modern portfolio theory~\cite{sharpe}, a Sharpe ratio of 1 is considered decent. About 2 or higher is very good and 3 is considered excellent. Table \ref{tab:sharpe_ratio} presents this data for our agents' policies.
\begin{center}
    \begin{table}[!htbp]
        \centering
        \caption{Sharpe Ratios for different approaches}
        \begin{tabular}{|c|c|c|c|}
            \hline
            Agent & Sharpe Ratio MSFT & Sharpe Ratio AMZN & Sharpe Ratio TSLA \\
            \hline
            Random Policy & -2.249 & -1.894 & -2.113  \\
            Without Sentiment & -1.357 & 1.487 & 0.926 \\
            \textbf{With Sentiment} & \textbf{2.432} & \textbf{2.212} & \textbf{1.874} \\
            \hline
        \end{tabular}
        \label{tab:sharpe_ratio}
    \end{table}
    
\end{center}
%\vspace{-0.2in}
The result for the random policy is as expected. It learns a terrible policy and its Sharpe ratio is the least among all three approaches. Surprisingly, the agent without sentiment data learns a pretty poor policy as well (albeit still better than the random policy), despite making profits. On closer analysis, it turns out that the MSFT stock had a general upward trend already and due to this reason a not-so-good policy could also produce profits, despite making sub-optimal decisions as indicated by the Sharpe ratio. Finally, we come to the agent which learnt a trading policy along with the sentiment data. Not only did it procure the highest profits as stated earlier, but also its decision making was very good as evidenced by its Sharpe ratio of 2.4 for MSFT, 2.2 for AMZN, and close to 2 for TSLA.

\section{Discussion and Conclusion} \label{sec:conclusion}
Much of the information about the real environment has been left out in this effort since we wanted to work from the ground up, looking just at how the sentiment data adds to the analysis. The daily closing price and the volume of data being traded for the last day or for the last "x-day" window (eg. 5-day, 50-day windows) could add further information to the environment as well.

Furthermore, instead of explicitly extracting the last "x-day" window opening price, we could use a Recurrent Neural Network for the network to retain on some historical trend information intrinsically. Initial experiments with an RNN proved difficult to optimize for the network, possibly due to noise in the data as well as probably not having the right hyper-parameters. This version of the network with RNNs took particularly long to train and was difficult to analyse because there was no way to extract what was happening in the hidden state of the network, so we took an alternative approach of explicitly providing it the last 5 and 50 day average opening price.

Also, our stock trading bot was limited to buying/selling a single stock per day, which very likely limits the amount of profit it could make. Making the agents action space 2-dimensional where the second dimension specifies the number of stocks bought/sold should be an easy way to remedy this. We tried an initial attempt at giving it the ability to trade with 1 stock or 5 stock per day, but the state space became much larger and coming up with a reward scheme that worked for this problem as well proved to be quite challenging.

In the real world, trading takes place at much higher frequencies than at an intra-day frequency; extending this to a much finer granular level with data on a second-by-second or minute-by-minute basis should be straightforward with our current framework. Also our work focuses on using stock data of a single company, but it can easily be extended to use stock data from multiple entities.

Also, in the knowledge graph we kept the relationship distance threshold quite limited so as to limit the noise added to the data in terms of news headlines. Provided with a knowledge graph which has weighted nodes, which tell if there is a positive or negative relationship between the entity in question and the company stocks are being traded for, we can potentially exploit much longer distance relationships and in a much more accurate manner.

%\section{Conclusion}

We present an approach of extracting implicit relationships between entities from news headlines via knowledge graphs and exploiting sentiment analysis, positive or negative, on these related headlines,  and then using this information, train a reinforcement learning agent. The trained reinforcement learning agent can perform better in terms of profits incurred as compared to an agent which does not have this additional information on headline sentiments. The whole pipeline as such is a novel approach and the empirical study demonstrates its validity.

\bibliographystyle{splncs04}
\bibliography{mybibliography}

\end{document}